\documentclass[lettersize,journal]{IEEEtran}
\usepackage{amsmath,amsfonts}
\usepackage{algorithmic}
\usepackage{algorithm}
\usepackage{array}
\usepackage[caption=false,font=normalsize,labelfont=sf,textfont=sf]{subfig}
\usepackage{textcomp}
\usepackage{stfloats}
\usepackage{url}
\usepackage{verbatim}
\usepackage{graphicx}
\usepackage{cite}

\usepackage{times}
\usepackage{epsfig}
\usepackage{amsmath}
\usepackage{amssymb}
\usepackage{eso-pic}

\usepackage{lineno}
\usepackage{booktabs,makecell, multirow, tabularx}
\usepackage{colortbl}  % 添加表格颜色支持
\usepackage{threeparttable}

\usepackage{tikz,xcolor}
\usepackage{hyperref}

\hypersetup{hidelinks,
	colorlinks=true,
	allcolors=black,
	pdfstartview=Fit,
	breaklinks=true}

\begin{document}

\title{DSKC: Domain Style Modeling with Adaptive Knowledge Consolidation for Exemplar-free Lifelong Person Re-Identification}

\author{
Shiben Liu \href{https://orcid.org/0000-0001-9376-2562}{\includegraphics[scale=0.08]{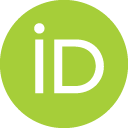}}, 
Mingyue Xu,
Huijie Fan
\href{https://orcid.org/0000-0002-8548-861X}{\includegraphics[scale=0.08]{ORCIDiD_icon128x128.png}, ~\IEEEmembership{Member, IEEE}}, 
Qiang Wang \href{https://orcid.org/0000-0002-2018-1764}{\includegraphics[scale=0.08]{ORCIDiD_icon128x128.png}}, 
Liangqiong Qu \href{https://orcid.org/0000-0001-8235-7852}{\includegraphics[scale=0.08]{ORCIDiD_icon128x128.png}, ~\IEEEmembership{Member, IEEE}},
Zhi Han \href{https://orcid.org/0000-0002-8039-6679}{\includegraphics[scale=0.08]{ORCIDiD_icon128x128.png}, ~\IEEEmembership{Member, IEEE}},
\thanks{This work is supported by the National Natural Science Foundation of China (62273339, U24A201397), and the LiaoNing Revitalization Talents Program (XLYC2403128). (\emph{Corresponding author: Huijie Fan})}
\thanks{Shiben Liu, and Mingyue Xu are with the State Key Laboratory of Robotics and Intelligent Systems, Shenyang Institute of Automation, Chinese Academy of Sciences, Shenyang 110016, China, and with the University of Chinese Academy of Sciences, Beijing 100049, China (e-mail: liushiben@sia.cn, xumingyue24@mails.ucas.ac.cn).
\par
Huijie Fan, Yandong Tang, and Zhi Han are with the State Key Laboratory of Robotics and Intelligent Systems, Shenyang Institute of Automation, Chinese Academy of Sciences, Shenyang 110016, China (e-mail: fanhuijie@sia.cn, ytang@sia.cn, hanzhi@sia.cn).
\par
Qiang Wang is with the Key Laboratory of Manufacturing Industrial Integrated Automation, Shenyang University, and with the State Key Laboratory of Robotics, Shenyang Institute of Automation, Chinese Academy of Sciences, Shenyang, 110016, China (e-mail: wangqiang@sia.cn). \\
\par
Liangqiong Qu is with the Department of Statistics and Actuarial Science and the Institute of Data Science, The University of Hong Kong, Hong Kong, 999077 (e-mail: liangqqu@hku.hk).\\
}
}
\markboth{Journal of \LaTeX\ Class Files,~Vol.~14, No.~8, August~2021}%
{Shell \MakeLowercase{\textit{et al.}}: A Sample Article Using IEEEtran.cls for IEEE Journals}
\maketitle
\begin{abstract}
Lifelong Person Re-identification (LReID) aims to continuously match individuals across camera views from sequential data streams. Existing LReID methods often ignore domain-specific style awareness and unified knowledge consolidation, which are crucial for mitigating forgetting when adapting to new information. We propose DSKC, a novel rehearsal-free and distillation-free framework for LReID. DSKC designs a domain-style encoder (DSE) to dynamically model domain-specific styles, and a unified knowledge consolidation (UKC) mechanism to adaptively integrate instance-level representations with domain-specific style into a cross-domain unified representation. By leveraging unified representation as a bridge, DSKC explicitly models inter-domain associations at both instance and domain levels to enhance anti-forgetting and generalization. Experimental results demonstrate that our DSKC outperforms state-of-the-art methods in two training orders and enhances the model's strong performance. Our code is available at \url{https://github.com/LiuShiBen/DKUA}. 
\end{abstract}
\begin{IEEEkeywords}
	Lifelong learning, Domain-style modeling, Exemplar-free, person re-identification.
\end{IEEEkeywords}
\begin{figure}[ht]
	%\vskip 0.2in
	\begin{center}
		\centerline{\includegraphics[width=\columnwidth]{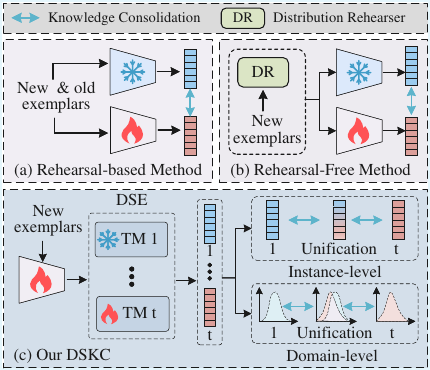}}
		\caption{
			(a) Rehearsal-based approaches enforce representation alignment by storing old samples, raising data privacy concerns. (b) Rehearsal-free methods only model the style of the last domain, limiting the model’s long-term anti-forgetting capacity. (c) Our DSKC, a novel rehearsal-free and distillation-free framework, dynamically models domain-specific styles and transforms current instances into different representations with dedicated domain-specific styles.
		}
		\label{fig-Introduction}
	\end{center}
\end{figure} 
\section{Introduction}
Person re-identification (ReID) aims to match individuals across non-overlapping camera views. Although existing ReID methods \cite{sun2025flexireid, lu2025llavareid} excel in static scenarios, they struggle to adapt to real-world environments where domain styles evolve continuously. This has motivated research into Lifelong Person Re-identification (LReID) \cite{pu2021lifelong, liu2025Diverse}, which requires models to incrementally acquire knowledge from sequential domains. The core challenge of LReID lies in balancing the preservation of previously learned knowledge with the adaptation to emerging information.  \\ 
Current LReID research primarily follows two branches: rehearsal-based \cite{pu2021lifelong, yu2023lifelong} and rehearsal-free methods \cite{xu2024distribution, qian2024auto}. As shown in Fig. \ref{fig-Introduction}(a), rehearsal-based schemes align representations via knowledge distillation using stored exemplars. However, these approaches entail privacy risks and high computational overhead \cite{zhao2021continual, luo2025lifelong}. Conversely, existing rehearsal-free solutions \cite{xu2025dask} often rely on modeling the style of the most recent domain for knowledge consolidation (Fig. \ref{fig-Introduction}(b)), which limits long-term anti-forgetting capacity. Despite these efforts, two critical issues remain: \textbf{Domain-specific Style Awareness}. Environmental variations (e.g., lighting, backgrounds) cause significant style shifts, leading to biased learning where models achieve the strong performance of current domain at the expense of historical ones \cite{liu2025Diverse}. We consider that dedicated modules to model domain-specific styles \cite{douillard2022dytox} without storing exemplars are essential to mitigate catastrophic forgetting. 2) \textbf{Unified Knowledge Consolidation}. Some methods \cite{sun2022patch, huang2023learning} typically align the current representations with instance-level outputs from the old model. However, this fails to ensure the acquisition of cross-domain unified knowledge, making it difficult to bridge inter-domain gaps. In general, integrating domain-specific awareness with unified knowledge learning is crucial for balancing old knowledge preservation and new information adaptation.\\
\indent To address these issues, we propose the domain style modeling with adaptive knowledge consolidation (DSKC) framework. As illustrated in Fig. \ref{fig-Introduction}(c), DSKC explores domain-specific representations with the dedicated domain styles and the unified representation served as cross-domain shared knowledge for enhancing model's anti-forgetting and adaptation capacity. Specifically, we design a domain-style encoder (DSE) to transfer instance-level representations of the current domain into the domain-specific representations for significantly preserving acquired knowledge. Subsequently, a unified knowledge learning (UKL) module adaptively integrates these representations into cross-domain unified representations, and enhance knowledge consolidation across domains. To further alleviate catastrophic forgetting, a unified knowledge association (UKA) module employs the unified representations as a bridge to explicitly model and align inter-domain relationships, thereby narrowing domain gaps. Finally, domain-based style transfer (DST) is introduced to enhance current domain-style modeling and facilitate cross-domain knowledge acquisition,  bolstering the model's performance in complex scenarios. Our contributions are summarized as follows:
\begin{itemize}
	\item We propose DSKC, a novel rehearsal-free and distillation-free framework for LReID. DSKC dynamically model domain-specific style and transforms the current instances into different representations with the dedicated domain style, effectively preserving learned knowledge without storing exemplars.
	\item We develop a UKL module, which adaptively integrates domain-specific representations into a unified representation space while enhancing knowledge consolidation across multiple domains.
	\item To effectively associate cross-domain information, the UKA is proposed, leveraging the unified representation as a bridge to explicitly maintain consistency among domain-specific representations for each instance.
	\item We design a DST mechanism to enhance current domain-style modeling and facilitate cross-domain knowledge acquisition, bolstering the model's robustness and performance in complex scenarios.
\end{itemize}
\section{Related work}
\subsection{Lifelong Person Re-Identification}
Lifelong person re-identification (LReID) suffers from a key challenge, which involves a trade-off between the preservation of old knowledge and adaptation to new information. To solve this challenge, existing LReID approaches \cite{wu2021generalising, xu2024mitigate, xu2025long, Cui2025Differentiated} mainly consist of rehearsal-based and rehearsal-free schemes. Rehearsal-based approaches  \cite{yu2023lifelong, liu2025Diverse} focus on knowledge distillation to achieve knowledge alignment by jointly training old and new exemplars. Despite their excellent performance in anti-forgetting ability, they inevitably raise privacy concerns and consume computational resources. Rehearsal-free solutions \cite{li2024exemplar, xu2024distribution} retain previous domain styles as old knowledge prior to preserve old knowledge forgetting, limiting the model's long-term anti-forgetting capacity. In this paper, to overcome these limitation, we propose a rehearsal-free and distillation-free framework to explore each domain-specific style and the unified knowledge of all domains for enhancing anti-forgetting and generalization capacity.
\subsection{Domain-style Awareness}
In lifelong learning, style-aware methods \cite{liang2024inflora} consolidate each domain-specific style to preserve the learned knowledge, rather than storing old exemplars. Some methods learn domain-specific knowledge by assigning the same structure, including the expert network \cite{yu2024boosting, liu2025domain} and the adapter \cite{qian2024auto}. Additionally, some works \cite{xu2024distribution} focus on modeling the data uncertainty to handle the out-of-distribution data for increasing the richness of data samples and alleviating old knowledge forgetting. In this paper, we investigate the dedicated style of each domain without storing historical exemplars to significantly preserve old knowledge and achieve consistency alignment at the instance and domain level.
\begin{figure*}[ht]
	%\vskip 0.2in
	\begin{center}
		\centerline{\includegraphics[width=1\linewidth, height=0.4\textheight]{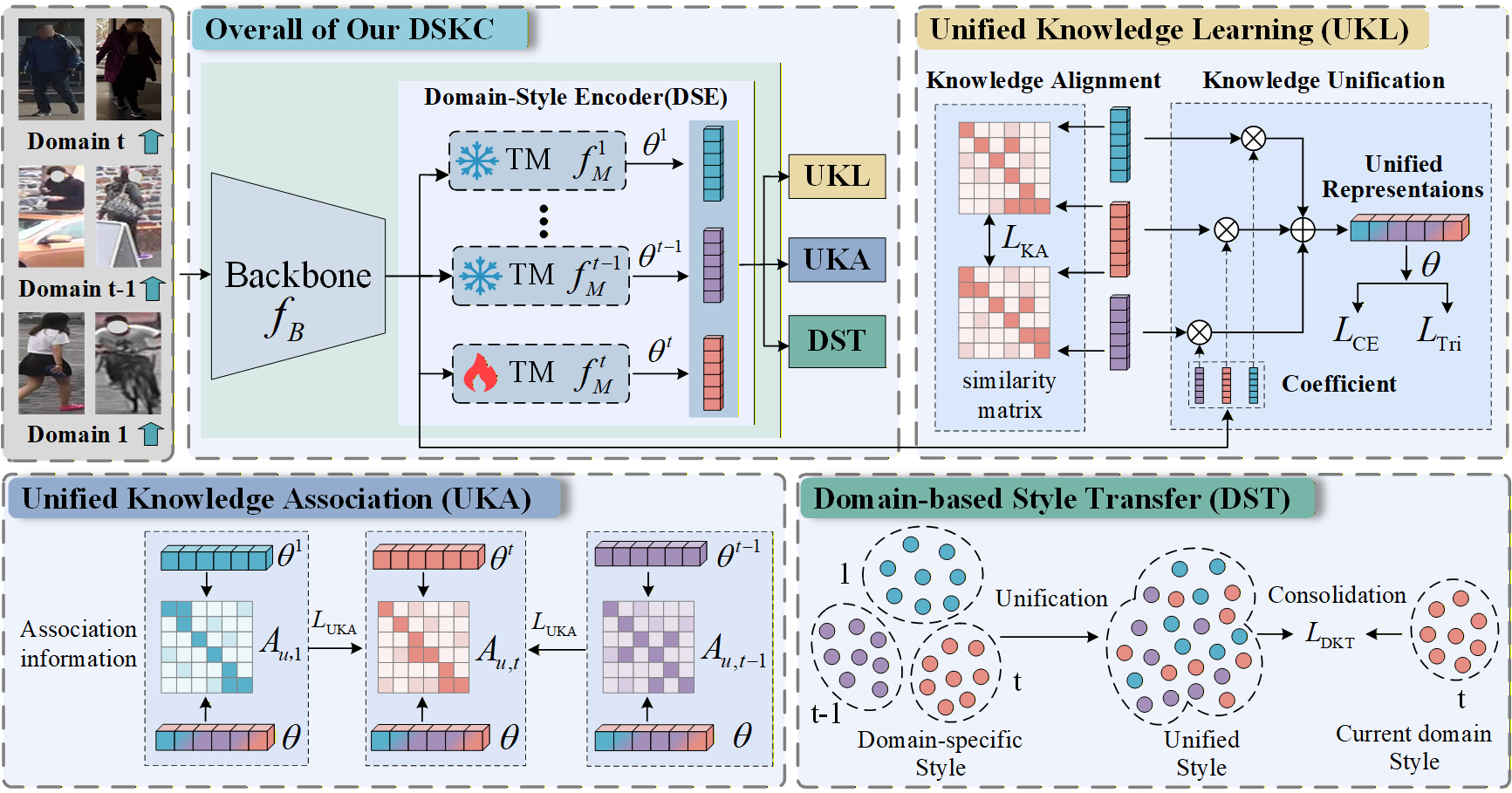}}
		\caption{Overview of our DSKC framework: the domain-style encoder transforms the robust representation of each instance to domain-specific representations $\{\theta^i\}_{i=1}^{t}$ at the $t$th domain. The UKL then dynamically integrates these representations into a unified representation $\theta$, and enhance knowledge consolidation across domains. The UKA leverages $\theta$ as a bridge to establish inter-domain associations $\{A_{u-i}\}_{i=1}^{t}$. Finally, the DST to enhance current domain-style modeling and facilitate cross-domain knowledge acquisition.}
		\label{fig:framework}
		\vspace{-25pt}
	\end{center}
\end{figure*}
\section{Methodology}
\subsection{Preliminary}
%$D^t=\{(x_i, y_i)\}_{i=1}^N_t$
In the rehearsal-free LReID task, a sequence of $T$ training domains $D=\{D^t\}_{t=1}^T$ is provided for successive model training. Each domain $D^t =\{(x_i, y_i)\}_{i=1}^{N_t}$ contains $N_t$ images $x_i$ and corresponding identity labels $y_i$. During the $t$-th domain, the model can only access the current domain $D^t$ with no availability of previous domain $\{D^i\}_{i=1}^{t-1}$. The architecture comprises a Backbone $f_B$, a domain-style encoder (DSE) with $t$-1 frozen transfer module (TM) $\{f_M^i\}_{i=1}^{t-1}$ and an updatable TM $f_M^t$ for the current domain. To evaluate the model’s anti-forgetting and generalization ability, the model is tested the Seen domain and Unseen domain.
\subsection{Overview of DKUA}
%DSKC: Domain Style Modeling with Adaptive Knowledge Consolidation for Exemplar-free Lifelong Person Re-Identificatio
We propose a novel domain style modeling with adaptive knowledge consolidation (DSKC) with a rehearsal-free and distillation-free framework, as illustrated in Figure. \ref{fig:framework}. Our approach consists of four key components: 1) Domain-style encoder, 2) unified knowledge learning (UKL), 3) unified knowledge association (UKA), and 4) Domain-based style transfer (DST). 
\subsection{Domain-style encoder}
To effectively preserve knowledge, existing methods \cite{liu2025Diverse,Cui2025Differentiated} often rely on storing exemplars, which raises data privacy concerns. In contrast, we aim to learning domain-specific representations that capture unique domain styles to retain knowledge without storing raw samples. Specifically, we propose a domain-style encoder (DSE) to transform instance-level features of the current domain into various domain-specific styles to alleviate catastrophic forgetting, as illustrated in Figure. \ref{fig:framework} (Domain-Style Encoder). \\ 
%Given a batch input image $x\in\mathbb{R}^{B\times C\times H\times W}$, the Backbone $f_B$ employs a vision transformer (ViT) with a depth setting of eleven to extract robust representations $f_B(x)\in\mathbb{R}^{B\times D}$ for each instance.
\indent As noted in \cite{douillard2022dytox}, dynamic architectures can effectively mitigate catastrophic forgetting. Inspired by this, we design a domain-style encoder comprising $T$ transfer module (TM). At training domain $t$, this architecture offers the following advantages: 1) the current TM $f_M^t$ adaptively captures knowledge from the new domain; 2) previous TMs $\{f_M^i\}_{i=1}^{t-1}$ are frozen to reliably retain knowledge from their respective historical domains; and 3) each instance from the current domain is transformed into $t$ domain-specific representations, capturing dedicated domain styles to preserve old knowledge without storing exemplars. Domain-specific representations $\{\theta^i\}_{i=1}^{t}$ generated by domain-style encoder are calculated as:  \\
\begin{equation}
	\theta^i = f_M^i(f_B(x))  (i=1,2,\cdots,t)
\end{equation}
where $f_M^i$ indicates transfer module (TM) operations including a multi-head self-attention (MHSA), feed-forward network (FFN), and a linear layer.
\subsection{Unified Knowledge Learning}
The domain-style encoder (DSE) captures domain-specific representations $\{\theta^i\}_{i=1}^{t}\in\mathbb{R}^{B\times D}$ with the dedicated domain style from the each instance of current domain $t$. To bridge inter-domain gaps, we aim to derive a unified representation by integrating these domain-specific representations in the latent space. To this end, we propose unified knowledge learning (UKL) to dynamically integrate domain-specific representations and enhance knowledge consolidation across multiple domains, as illustrated in Figure. \ref{fig:framework} (Unified Knowledge Learning). UKL comprises knowledge unification (KU) and knowledge alignment (KA).  \\
\textbf{Knowledge Unification:} We propose knowledge unification (KU) to dynamically integrate domain-specific representations $\{\theta^i\}_{i=1}^{t}\in\mathbb{R}^{B\times D}$ into a unified representation $\theta\in\mathbb{R}^{B\times D}$, which serves as a across-domain knowledge center. It effectively facilitates the learning of domain-invariant representations, thereby mitigating catastrophic forgetting. Specifically, we adaptively learn weighting coefficients $\omega \in \mathbb{R}^{B \times t}$ derived from robust representation $f_B(x)$ mappings, which can be formulated as:
\begin{equation}
	\omega = \delta(\phi(\mu(f_B(x))))
\end{equation}
where $\mu$($\cdot$), $\phi$($\cdot$), and $\delta$($\cdot$) denote the mean, linear, and softmax functions, respectively. The unified representation $\theta$ is then obtained via a weighted linear combination of domain-specific representations using the learned coefficients $\omega$,  which can be formulated as:
\begin{equation}
	\theta = \sum_{i=1}^{t} \omega^i\theta^i
\end{equation}
To ensure the new knowledge learning and the old knowledge anti-forgetting, we adopt the classical ReID loss including a cross-entropy loss $L_{\mathrm{CE}}$ and a triplet loss $L_{\mathrm{Tri}}$.
\begin{equation}
	L_{\mathrm{CE}} = -y\log(f_c(\theta))
\end{equation}
where $f_c$ is classifier and $y$ indicates identity label.
\begin{equation}
	L_{\mathrm{Tri}} = max(||\theta-\theta_p||_2^2-||\theta-\theta_n||_2^2 + m, 0)
\end{equation}
where $\theta_p$ and $\theta_n$ are the positive and negative samples related to an anchor sample $\theta$. $||\cdot||_2$ is $L$2 norm. The classical ReID loss is formulated as:
\begin{equation}
	L_{\mathrm{ReID}} = L_{\mathrm{CE}} + L_{\mathrm{Tri}}
\end{equation}
\textbf{Knowledge Alignment:} Significant gaps between training domains can introduce bias during knowledge unification, shifting the unified representation away from the true cross-domain center. To mitigate this, we propose knowledge alignment (KA) mechanism to refine knowledge consolidation across multiple domains. At the training domain $t$, we stepwise constrain the current representation $\theta^t$ against previous representations  $\{\theta^i\}_{i=1}^{t-1}$ to ensure the unified representation remains centered. Specifically, we compute the cosine similarity matrix $S_{t,i}$ between $\theta^t$ and each preceding domain-specific representation $\theta^i$. These matrices are then converted into cosine distance metrics, $1-S_{t,i}$, to enforce distributional convergence between the current and all historical domains. The KA mechanism is calculated as:
\begin{equation}
	S_{t, i} = Cos(\theta^{t}, \theta^{i}) (i=1,2,\cdots,t-1)
\end{equation}
\begin{equation}
	L_{\mathrm{KA}} = \frac{1}{t-1}\sum\limits_{i=1}^{t-1} \mu(1-S_{t, i})
\end{equation}
where $Cos(\cdot,\cdot)$ and $\mu$($\cdot$) are cosine similarity and mean function, respectively. 
\subsection{Unified Knowledge Association}
At the training domain $t$, the domain-specific and unified representations must satisfy two conditions: 1) each domain-specific representation $\theta^i$ should establish an affinity association with the unified representation $\theta$ to measure similarity across different domain styles; 2) the unified representation $\theta$ should serve as a bridge to pull various domain-specific representations $\{\theta^i\}_{i=1}^{t}$ closer, thereby enhancing domain-invariant learning. Accordingly, we propose the unified knowledge association (UKA) mechanism to align heterogeneous domain-specific representations and facilitate unified representation learning, as shown in Figure. \ref{fig:framework} (Unified Knowledge Association). Specifically, we define association information $\{A_{u,i}\}_{i=1}^{t}$ to estimate the relative distances between $\theta$ and each $\theta^i$, formulated as:
\begin{equation}
	A_{u,i} = \delta(Cos(\theta^t, \theta)/\lambda) (i=1,2,\cdots,t-1)
\end{equation}
where $\delta$ denotes the softmax function and $\lambda$ is a temperature parameter set to 0.1 \cite{hirzer2011person}. To refine the unified representation and narrow inter-domain gaps, we optimize the association information by minimizing the discrepancy between the current domain knowledge and the accumulated historical knowledge, formulated as:
\begin{equation}
	L_{\mathrm{UKA}} = \sum\limits_{i=1}^{t-1} KL(A_{u,t}, A_{u,i})
\end{equation}
where $KL(\cdot,\cdot)$ is kullback leibler divergence \cite{hershey2007approximating}.\\
\subsection{Domain-based style transfer}
While UKL and UKA primarily operate at the instance level, we further introduce a Distribution-based Knowledge Transfer (DKT) mechanism to enhance global performance. DKT prevents the current domain distribution from deviating from the unified distribution, thereby bolstering anti-forgetting and adaptation capabilities. Specifically, we represent each domain's distribution using a covariance matrix derived from all instances within that domain, formulated as:
\begin{equation}
	\Sigma^t = \frac{1}{N_t-1} \sum_{j=1}^{N_t} (\theta_j^t - \bar{\theta_j^t})(\theta_j^t - \bar{\theta_j^t})^\top
\end{equation}
where $\theta_j^t$ and $\bar{\theta}^t$ denote the domain-specific representations and their corresponding mean across all instances in the current domain. Similarly, we derive the covariance matrices $\{\Sigma^{i}\}_{i=1}^{t}$ for all domains. Subsequently, these individual distributions are integrated into a unified covariance matrix $\Sigma^{1,t}$, representing the cross-domain distribution according to the following scheme:
\begin{equation}
	\Sigma^{1,t} = \Sigma^{1,t-1} \cdot \frac{|y^{1,t-1}|}{|y^{1,t}|} + \Sigma^t \cdot \frac{|y^{1,t}| - |y^{1,t-1}|}{|y^{1,t}|}
\end{equation}
where $y^{1,t-1}$ and $y^{1,t}$ denote the cumulative number of classes in the first $t-1$ and $t$ domains, respectively. Finally, we minimize the discrepancy between the unified and current domain distributions using the Kullback-Leibler (KL) divergence \cite{hershey2007approximating}, formulated as:
\begin{equation}
	L_{\mathrm{jdc}} = KL(\Sigma^{1,t},\Sigma^{t})
\end{equation}
The overall loss function $L$ of our method is formulated as:
\begin{equation}
	L = L_{\mathrm{ReID}} + L_{\mathrm{KA}} + L_{\mathrm{UKA}} + L_{\mathrm{DKT}}
\end{equation}
\begin{table}[t]
	\centering
	\renewcommand{\arraystretch}{1.3}
	\setlength{\tabcolsep}{5pt}
	\caption{Dataset statistics of the LReID benchmark. Since the sampling procedure results in the numbers of train IDs being all 500, the original numbers of IDs are listed for comparison. '-' denotes that the dataset is not used for training} \label{tab:Table1}
	\begin{tabular}{l|l|l|c|c}  %c:center r:right l:left
		\hline
		Type &Datasets &Scale &Train IDs &Test IDs \cr
		\hline
		\multirow{5}{*}{Seen}&
		Market-1501\cite{zheng2015scalable}  &Large &500 (751) &750 \cr
		&CUHK-SYSU\cite{xiao2016end}  &Mid &500 (942) &2900 \cr
		&DukeMTMC\cite{ristani2016performance}  &Large &500 (702) &1110 \cr
		&MSMT17$\_$V2\cite{wei2018person} &Large &500 (1041) &3060 \cr
		&CUHK03\cite{li2014deepreid} &Mid &500 (700) &700 \cr
		\hline
		\multirow{6}{*}{Unseen}&
		VIPeR\cite{gray2008viewpoint} &Small &\makecell[c]{$-$} &316  \cr
		&GRID\cite{loy2010time} &Small &\makecell[c]{$-$} &126 \cr
		&CUHK02\cite{li2013locally} &Mid &\makecell[c]{$-$} &239 \cr
		&Occ$\_$Duke\cite{miao2019pose} &Large &\makecell[c]{$-$} &1100 \cr
		&Occ$\_$REID\cite{zhuo2018occluded} &Mid &\makecell[c]{$-$} &200 \cr
		&PRID2011\cite{hirzer2011person} &Small &\makecell[c]{$-$} &649 \cr
		\hline
	\end{tabular}
\end{table}
\section{Experiments}
\subsection{Experiments Setting}
\textbf{Datasets:} All experiments are conducted on a challenging LReID benchmark \cite{pu2021lifelong}, which comprises Seen and Unseen domains. The Seen domain includes Market-1501\cite{zheng2015scalable}, CUHK-SYSU \cite{xiao2016end}, DukeMTMC \cite{ristani2016performance}, MSMT17$\_$V2 \cite{wei2018person} and CUHK03 \cite{li2014deepreid}. The Unseen domain consists of VIPeR \cite{gray2008viewpoint}, GRID \cite{loy2010time}, CUHK02 \cite{li2013locally}, Occ$\_$Duke \cite{miao2019pose}, Occ$\_$REID \cite{zhuo2018occluded}, and PRID2011 \cite{hirzer2011person}. Following \cite{pu2021lifelong}, we employ two classic training orders to evaluate model performance. These Unseen domain incorporate variations in shape, lighting conditions, and occlusion to effectively demonstrate the generalization capacity of our method. Dataset statistics of the LReID benchmark are shown in Table \ref{tab:Table1}. \\
\begin{table*}[t]
	\renewcommand{\arraystretch}{1.3}
	\setlength{\tabcolsep}{5pt}
	\caption{Performance comparison with state-of-the-art methods on training order-1. Bolded black and red fonts denotes the optimal and suboptimal values, respectively. Training order-1 is Market-1501$\to$CUHK-SYSU$\to$ DukeMTMC$\to$MSMT17$\_$V2$\to$CUHK03.}
	\label{tab:Table2}
	\begin{center}
		\begin{small}
			\begin{sc}
				\begin{tabular}{l|cc|cc|cc|cc|cc|cc|cc}
					\toprule
					\multirow{2}{*}{Method}&
					\multicolumn{2}{c|}{Market-1501}&\multicolumn{2}{c|}{CUHK-SYSU}&\multicolumn{2}{c|}{DukeMTMC}&\multicolumn{2}{c|}{MSMT17$\_$V2}&
					\multicolumn{2}{c|}{CUHK03}&\multicolumn{2}{c}{\textbf{Seen-Avg}}&\multicolumn{2}{c}{\textbf{Unseen-Avg}}\\
					\cline{2-15}
					&mAP &R@1 &mAP &R@1 &mAP &R@1 &mAP &R@1 &mAP &R@1 &mAP &R@1 &mAP &R@1\cr
					\midrule
					AKA &58.1 &77.4 &72.5 &74.8 &28.7 &45.2 &6.1 &16.2 &38.7 &40.4 &40.8 &50.8 &42.0 &39.8 \\
					PTKP &64.4 &82.8 &79.8 &81.9 &45.6 &63.4 &10.4 &25.9 &42.5 &42.9 &48.5 &59.4 &51.2 &49.1 \\
					PatchKD &68.5 &85.7 &75.6 &78.6 &33.8 &50.4 &6.5 &17.0 &34.1 &36.8 &43.7 &53.7 &45.1 &43.3\\
					KRKC &54.0 &77.7 &83.4 &85.4 &48.9 & 65.5 &14.1 &33.7 &49.9 &50.4 &50.1 &62.5 &52.7 &50.8\\
					C2R &69.0 &86.8 &76.7 &79.5 &33.2 &48.6 &6.6 &17.4 &35.6 &36.2 &44.2 &53.7 &- &-\\
					LSTKC &54.7 &76.0 &81.1 &83.4 &49.4 &66.2 &20.0 &43.2 &44.7 &46.5 &50.0 &63.1 &51.3 &48.9\\
					DRE &69.4 &85.2 &85.6 &86.8 &52.8 &67.3 &21.5 &43.7 &54.7 &58.1 &56.8 &68.2 &56.7 &55.0 \\
					\midrule
					DKP &60.3 &80.6 &83.6 &85.4 &51.6 &68.4 &19.7 &41.8 &43.6 &44.2 &51.8 &64.1 &49.9 &46.4\\
					DASK &61.2 &82.3 &81.9 &83.7 &58.5 &74.6 &\textcolor{red}{\textbf{29.1}} &\textbf{57.6} &46.2 &48.1 &55.4 &69.3 &56.3 &52.6\\
					PAEMA &71.7 &86.9 &\textcolor{red}{\textbf{90.0}} &\textcolor{red}{\textbf{91.3}} &\textbf{66.0} &\textbf{79.0} &\textbf{31.4} &\textcolor{red}{\textbf{55.3}} &50.0 &50.8 &61.8 &72.7 &61.1 & 57.9 \\
					DCR &75.9 &87.9 &87.3 &88.5 &60.1 &71.9 &25.3 &50.1 &60.5 &61.3 &61.8 &71.9 &60.8 &58.3\\
					\midrule
					DSKC-IM &\textcolor{red}{\textbf{80.3}} &\textcolor{red}{\textbf{89.7}} &87.5 &88.6 &58.7 &73.4 &27.0 &50.6 &\textbf{63.1} &\textbf{64.8}	&\textcolor{red}{\textbf{63.3}} &\textcolor{red}{\textbf{73.4}} &\textcolor{red}{\textbf{62.9}} &\textcolor{red}{\textbf{60.5}} \\
					Ours &\textbf{82.5} &\textbf{91.3} &\textbf{90.1} &\textbf{91.5} &\textcolor{red}{\textbf{60.9}} &\textcolor{red}{\textbf{75.4}} &25.9 &51.4 &\textcolor{red}{\textbf{61.2}} &\textcolor{red}{\textbf{62.8}} &\textbf{64.1} &\textbf{74.5} &\textbf{64.7} &\textbf{61.3} \\							
					\bottomrule
				\end{tabular}
			\end{sc}
		\end{small}
	\end{center}
\end{table*}
\begin{table*}[t]
	\renewcommand{\arraystretch}{1.3}
	\setlength{\tabcolsep}{5pt}
	\caption{Performance comparison with state-of-the-art methods on training order-2. Bolded black and red fonts denotes the optimal and suboptimal values, respectively. Training order-2 is DukeMTMC$\to$MSMT17$\_$V2$\to$Market-1501$\to$ CUHK-SYSU$\to$ CUHK03.}
	\label{tab:Table3}
	\begin{center}
		\begin{small}
			\begin{sc}
				\begin{tabular}{l|cc|cc|cc|cc|cc|cc|cc}
					\toprule
					\multirow{2}{*}{Method}			&\multicolumn{2}{c|}{DukeMTMC}&\multicolumn{2}{c|}{MSMT17$\_$V2}&\multicolumn{2}{c|}{Market-1501}&\multicolumn{2}{c|}{CUHK-SYSU}& \multicolumn{2}{c|}{CUHK03}&\multicolumn{2}{c}{\textbf{Seen-Avg}}&\multicolumn{2}{c}{\textbf{Unseen-Avg}}\\
					\cline{2-15}
					&mAP &R@1 &mAP &R@1 &mAP &R@1 &mAP &R@1 &mAP &R@1 &mAP &R@1 &mAP &R@1\cr 
					\midrule
					AKA &42.2 &60.1 &5.4 &15.1 &37.2 &59.8 &71.2 &73.9 &36.9 &37.9 &38.6 &49.4 &41.3 &39.0\\
					PTKP &54.8 &70.2 &10.3 &23.3 &59.4 &79.6 &80.9 &82.8 &41.6 &42.9 &49.4 &59.8 &50.8 &48.2\\
					PatchKD &58.3 &74.1 &6.4 &17.4 &43.2 &67.4 &74.5 &76.9 &33.7 &34.8 &43.2 &54.1 &44.8 &43.3\\
					KRKC &50.6 &65.6 &13.6 &27.4 &56.2 &77.4 &83.5 &85.9 &46.7 &46.6 &50.1 &61.0 &52.1 &47.7\\
					LSTKC &49.9 &67.6 &14.6 &34.0 &55.1 &76.7 &82.3 &83.8 &46.3 &48.1 &49.6 &62.1 &51.7 &49.5\\		
					C2R &59.7 &75.0 &7.3 &19.2 &42.4 &66.5 &76.0 &77.8 &37.8 &39.3 &44.7 &55.6 &- &-\\
					DRE &59.7 &74.2 &18.7 &34.8 &65.4 &82.7 &84.8 &86.7 &51.9 &53.2 &56.1 &66.3 &57.1 &54.9\\
					\midrule
					DKP &53.4 &70.5 &14.5 &33.3 &60.6 &81.0 &83.0 &84.9 &45.0 &46.1 &51.3 &63.2 &51.3 &47.8 \\
					DASK &55.7 &74.4 &25.2 &51.9 &\textcolor{red}{\textbf{71.6}} &\textcolor{red}{\textbf{87.7}} &84.8 &86.2 &48.4 &49.8 &57.1 &70.0 &57.7 &53.9\\
					PAEMA &\textcolor{red}{\textbf{67.2}} &\textcolor{red}{\textbf{79.8}} &26.0 &49.4 &69.8 &85.8 &\textbf{89.9} &\textbf{91.0} &49.3 &49.7 &60.4 &71.1 &60.5 &57.6\\
					DCR  &64.1 &77.2 &25.4 &44.9 &70.6 &84.5 &86.1 &\textcolor{red}{\textbf{88.2}} &54.2 &58.7 &60.1 &70.7 &61.6 &59.2\\
					\midrule
					DSKC-IM &66.1 &79.2 &\textcolor{red}{\textbf{29.6}} &\textcolor{red}{\textbf{54.6}} &68.5 &83.6 &86.4 &87.9 &\textbf{61.8} &\textbf{63.4} &\textcolor{red}{\textbf{62.5}} &\textcolor{red}{\textbf{73.7}} &\textcolor{red}{\textbf{64.2}} &\textcolor{red}{\textbf{61.4}}\\
					Ours &\textbf{69.5} &\textbf{81.8} &\textbf{31.8} &\textbf{57.6} &\textbf{73.7} &\textbf{87.9} &\textcolor{red}{\textbf{87.6}} &88.1 &\textcolor{red}{\textbf{59.5}} &\textcolor{red}{\textbf{60.9}} &\textbf{64.4} &\textbf{75.3} &\textbf{65.7} &\textbf{62.6}\\
					\bottomrule
				\end{tabular}
			\end{sc}
		\end{small}
	\end{center}
\end{table*}
\textbf{Implementation Details:} We adopt the ViT-Small/16 \cite{dosovitskiy2020image} architecture as our backbone, initialized with weights pre-trained on the large-scale unlabeled LUPerson dataset \cite{fu2021unsupervised}. Our DSKC framework is trained for 60 epochs per domain using the Adam optimizer \cite{kingma2014adam}. The batch size is set to 128, with an initial learning rate of $5 \times 10^{-6}$ that decays by a factor of 0.1 every 20 epochs. Our experiments are conducted on an NVIDIA RTX 6000 GPU.\\ 
\textbf{Evaluation Metrics.} Model performance is evaluated using two standard metrics: Mean Average Precision (mAP) \cite{zheng2015scalable} and Rank-1 accuracy (R@1) \cite{moon2001computational}. Furthermore, to assess lifelong learning ability, we report the average mAP and R@1 across all Seen and Unseen domains, providing a holistic measure of the model's generalization and anti-forgetting capabilities. \\
\subsection{Compared Methods} 
We compare our proposed DSKC against two categories of methods: rehearsal-based and rehearsal-free. Rehearsal-based approaches consist of AKA\cite{pu2021lifelong}, PTKP\cite{ge2022lifelong}, PatchKD\cite{sun2022patch}, KRKC\cite{yu2023lifelong},  LSTKC\cite{xu2024lstkc}, C2R\cite{cui2024learning}, and Vit-based DRE\cite{liu2025Diverse}. Rehearsal-free approaches include ResNet-based DKP\cite{xu2024distribution} and DASK\cite{xu2025dask}, Vit-based PAEMA\cite{li2024exemplar} and DCR \cite{liu2025domain}. DSKC-IM utilizes Vit-B/16 \cite{dosovitskiy2020image} backbone, initialized with ImageNet-21K \cite{he2021transreid}. Experimental results for training order-1 and training order-2 are presented in Table \ref{tab:Table2} and Table \ref{tab:Table3}, respectively.\\
\subsection{Performance Assessment on Seen domain} 
\textbf{Compared to Rehearsal-based Methods:} Rehearsal-based methods typically utilize knowledge distillation by jointly training on old exemplars and new data. As shown in Tables \ref{tab:Table2} and \ref{tab:Table3}, our DSKC significantly outperforms ResNet-based methods across all five domains, achieving Seen-Avg gains of \textbf{14.0\%}/\textbf{12.0\%} and \textbf{14.3\%}/\textbf{13.2\%} (mAP/R@1) under the two training orders, respectively. Furthermore, our DSKC outperforms the ViT-based DRE by \textbf{7.3\%}/\textbf{6.3\%} and \textbf{8.3\%}/\textbf{9.0\%} in Seen-Avg. In summary, by establishing affinity associations between unified and domain-specific knowledge at both instance and domain levels, our DSKC effectively narrows inter-domain gaps while maintaining an optimal balance between anti-forgetting and adaptation.\\
\textbf{Compared to Rehearsal-free Methods:} 
Rehearsal-free solutions typically retain domain styles as a proxy for historical knowledge. As shown in Tables \ref{tab:Table2} and \ref{tab:Table3}, our DSKC significantly outperforms ResNet-based DKP and DASK, achieving Seen-Avg gains of \textbf{8.7\%}/\textbf{5.2\%} and \textbf{7.3\%}/\textbf{5.3\%} (mAP/R@1) under the two training orders, respectively. Notably, our DSKC surpasses DASK on four out of five seen datasets. Although DASK performs better on MSMT17\_V2, due to its distribution-rehearsing mechanism and distillation-based alignment. Our DSKC still yields superior Seen-Avg improvements of \textbf{2.3\%}/\textbf{1.8\%} and \textbf{4.0\%}/\textbf{4.2\%} (mAP/R@1) compared to ViT-based PAEMA and DCR. 
These improvements stem from our domain-style encoder design, which effectively captures domain-specific styles without relying on stored exemplars, thereby achieving an optimal balance between anti-forgetting and adaptive learning.\\
\begin{figure}[t]
	\centering 
	\includegraphics[width=1\linewidth, height=0.16\textheight]{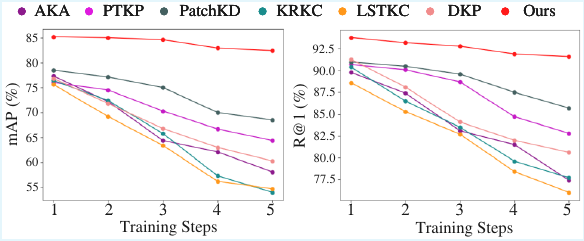}
	\caption{Anti-forgetting curves on Market-1501.}
	\label{fig:anti-forgetting}
\end{figure}
\textbf{Anti-forgetting Curves:} To evaluate anti-forgetting capabilities, we analyze performance on Market-1501 under Training Order-1, as illustrated in Figure. \ref{fig:anti-forgetting}. The results show that competing methods exhibit significant performance degradation after training on the large-scale MSMT17\_V2 dataset (Step 4). In contrast, our DSKC maintains remarkably stable mAP and R@1 scores as training progresses. This stability demonstrates that our DSKC effectively captures domain-specific knowledge to preserve historical style while facilitating unified representation learning, thereby significantly mitigating catastrophic forgetting.\\
\subsection{Performance Assessment on Unseen domain} 
\textbf{Compared to Rehearsal-based Methods:} 
As shown by the Unseen-Avg metrics in Tables \ref{tab:Table2} and \ref{tab:Table3}, our DSKC significantly outperforms rehearsal-based methods on the unseen domains. Specifically, it achieves incremental Unseen-Avg gains of \textbf{12.0\%}/\textbf{10.5\%} and \textbf{12.4\%}/\textbf{13.1\%} (mAP/R@1) across the two training orders, respectively. We attribute the performance limitations of ResNet-based models to two primary factors: 1) the constrained representation capacity of the architecture makes it challenging to distinguish between identities with high intra-class similarity; and 2) these models insufficiently consolidate unified cross-domain knowledge, leading to a degradation in generalization performance on unseen domains. Furthermore, compared to the ViT-based DRE method, our DSKC achieves superior performance, surpassing it by \textbf{7.3\%}/\textbf{6.3\%} and \textbf{8.3\%}/\textbf{9.0\%} (mAP/R@1) in Seen-Avg across the two training orders. Overall, our unified knowledge learning (UKL) design effectively leverages domain-specific representations for each instance to mine a unified representation across all domains, thereby significantly enhancing the model's generalization capabilities on unseen domains. \\
\textbf{Compared to Rehearsal-free Methods:} As presented in Tables \ref{tab:Table2} and \ref{tab:Table3}, our DSKC achieves superior performance, obtaining improvements of \textbf{3.6\%}/\textbf{3.0\%} and \textbf{4.1\%}/\textbf{3.4\%} (mAP/R@1) in Seen-Avg across the two training orders compared to the ViT-based PAEMA. This success can be attributed to our domain-style encoder design, which effectively maintains learned domain-specific styles at each training domain. By transforming instances from the current domain into various domain-specific styles, the model is guided to simultaneously capture unified cross-domain knowledge and identity-discriminative information. Consequently, the domain-specific representation learning significantly bolsters the model’s generalization capability across diverse domains.\\
\begin{figure}[t]
	\centering 
	\includegraphics[width=1\linewidth, height=0.15\textheight]{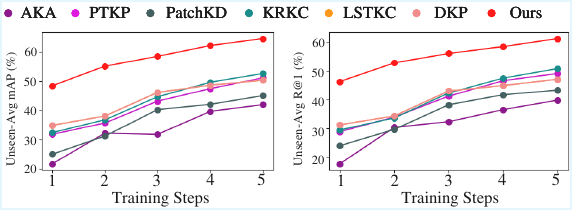}
	\caption{Generalization tendency on Unseen domain.}
	\label{fig:generalization}
\end{figure}
\textbf{Generalization of Unseen Domain:} As illustrated in Figure. \ref{fig:generalization}, compared to existing methods, our DSKC achieves superior performance growth on the unseen domains across all training steps, notably without utilizing knowledge distillation strategies or storing historical exemplars. In summary, our DSKC effectively achieves knowledge consistency alignment between domain-specific representations and the unified representation. This process facilitates the capture of domain-invariant knowledge, thereby significantly enhancing the model's generalization capabilities across diverse environments.
\begin{figure*}[ht]
	\begin{center}
		\includegraphics[width=1.0\linewidth,height=0.20\textheight]{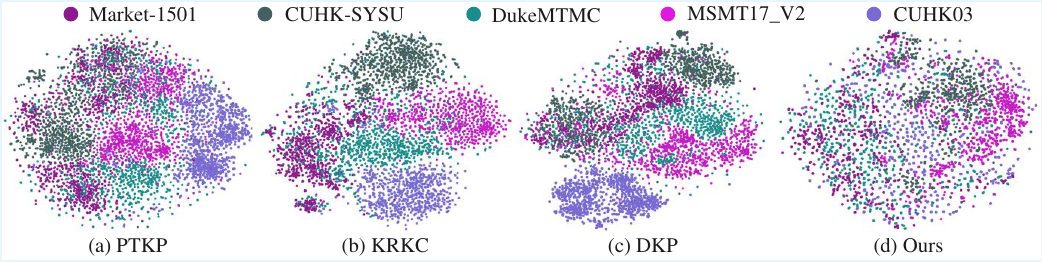}
		\caption{The t-SNE visualization results of the unified representation across five domains.}
		\label{fig:t-sne}
	\end{center}
\end{figure*}
\begin{table}[ht]
	\centering
	\renewcommand{\arraystretch}{1.3}
	\setlength{\tabcolsep}{5pt}
	\caption{Ablation studies of individual components on training order-1.}\label{tab:component}
	\begin{tabular}{cccc|c|c|c|c}
		\hline
		\multirow{2}{*}{Baseline} &\multirow{2}{*}{AKC} &\multirow{2}{*}{UKA} &\multirow{2}{*}{DKT} &\multicolumn{2}{c|}{Seen$\_$Avg} &\multicolumn{2}{c}{Unseen$\_$Avg}\\
		%\cmidrule(lr){2-4} \cmidrule(lr){5-7}\cmidrule(lr){8-10}
		\cline{5-8}
		&&&&mAP &R-1 &mAP &R-1 \cr \hline
		$\surd$ &&&&58.2 &65.8 &57.5 &54.6 \\
		$\surd$ &$\surd$ &&&59.5 &67.3 &58.9 &55.8 \\
		$\surd$ &$\surd$ &$\surd$ &&62.6 &72.1 &62.4 &59.7  \\
		$\surd$ &$\surd$ &$\surd$ &$\surd$ &\textbf{64.1} &\textbf{74.5} &\textbf{64.7} &\textbf{61.3}  \\ 
		\hline
	\end{tabular}
\end{table}
\subsection{Ablation Studies}
To assess the individual contributions of the components within our DSKC framework, we conduct extensive ablation studies to evaluate the impact of the Baseline, unified knowledge learning (UKL), unified knowledge association (UKA), Domain-based style transfer (DST) on overall model performance. The Baseline is composed of the domain-style encoder and the knowledge unification mechanism. These experiments are performed across both Seen and Unseen domains under training order-1, as illustrated in Table \ref{tab:component}. \\
\textbf{Performance of Different Components:} 
As shown in Table \ref{tab:component}, the Baseline validates that our domain-style encoder and knowledge unification strategy effectively capture domain-specific representations and consolidate the unified representation to mitigate catastrophic forgetting. Incorporating UKL (Row 2) further narrows inter-domain gaps, ensuring accurate and effective unified representation learning. The addition of the UKA mechanism (Row 3) yields significant improvements of \textbf{3.1\%}/\textbf{4.8\%} and \textbf{3.5\%}/\textbf{4.1\%} (mAP/R@1), demonstrating its efficacy in associating instance-level knowledge to enhance both anti-forgetting and generalization. Furthermore, the DST design (Row 4) enables the model to align domain-specific style into unified style across domains. Each component plays a vital role; when integrated, the full DSKC framework achieves the optimal balance between stability and plasticity.\\
\textbf{Performance of Unified Representation:} 
To validate the unified representation learning capability of our approach, we present the t-SNE visualization results of PTKP, KRKC, DKP, and our DSKC across five domains, as illustrated in Fig. \ref{fig:t-sne}. The visualization reveals that both KRKC and DKP struggle to accumulate sufficient cross-domain unified knowledge, which inherently limits their anti-forgetting capacity. Meanwhile, PTKP fails to adequately separate identity information within individual domains, leading to a noticeable decrease in matching accuracy. Compared to these existing methods, our DSKC exhibits two distinct advantages: 1) Comprehensive cross-domain interaction: The visualization demonstrates that our DSKC effectively exploits a unified representation across all domains, thereby significantly reducing inter-domain gaps. 2) Enhanced identity discriminability: The clearly differentiated identity clusters reveal that our DSKC dynamically consolidates domain-specific representations capable of effectively distinguishing similar identities across all domains. In summary, our DSKC successfully consolidates a unified representation that balances cross-domain commonality with identity discrimination, leading to superior anti-forgetting performance and matching accuracy.\\
\begin{figure*}[ht]
	%\vskip 0.2in
	\begin{center}
		\includegraphics[width=1.0\linewidth, height=0.36\textheight]{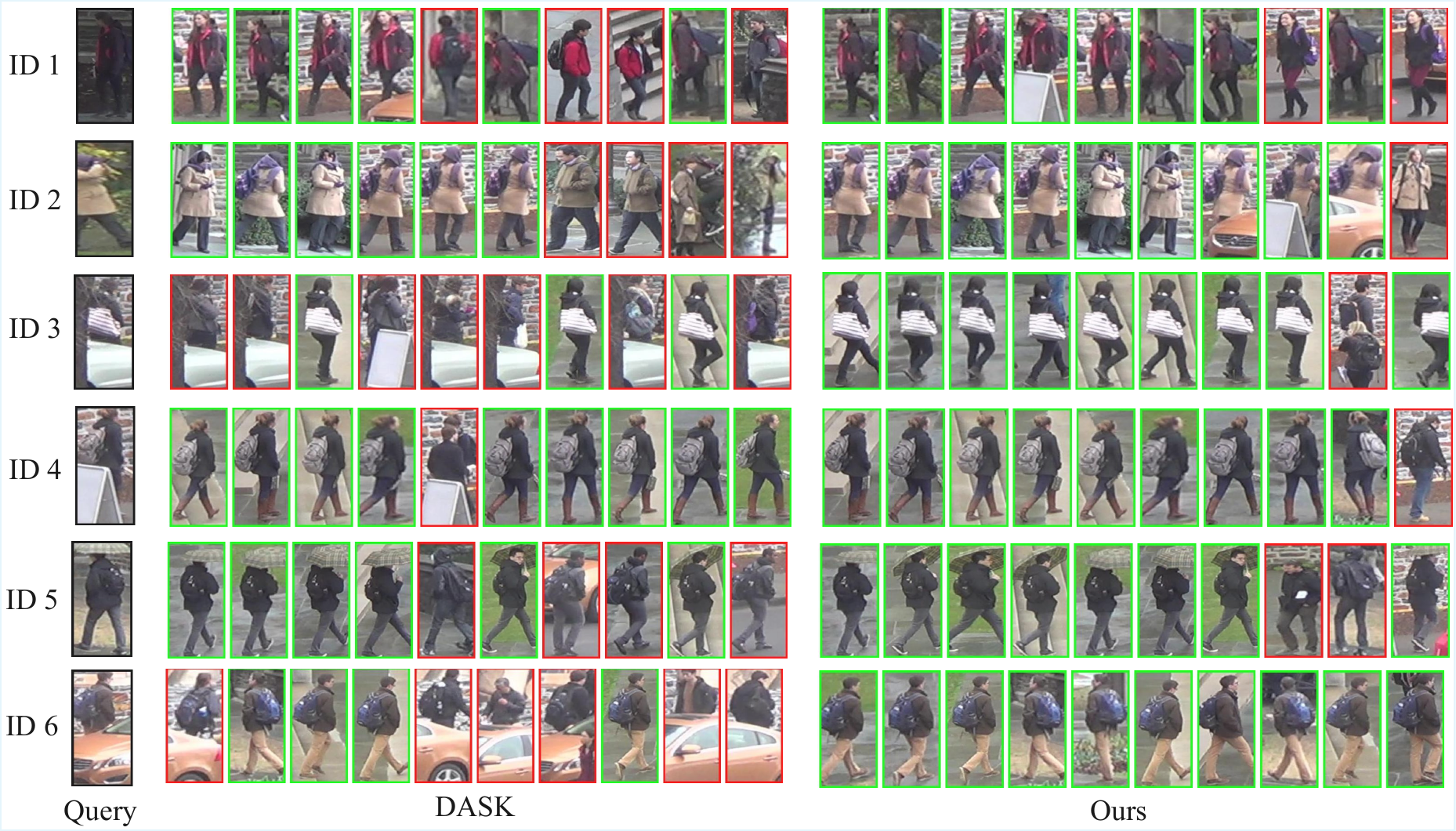}
		\caption{Visualization of retrieval results. In the retrieved results, correctly matched pedestrians are marked with green boxes, while incorrect matches are highlighted in red box.}
		\label{fig:visual}
	\end{center}
\end{figure*}
\subsection{Analysis of Computational Complexity} 
\textbf{Complexity Analysis of transfer module (TM): } Each domain is equipped with a dedicated Transfer Module (TM) designed to capture its domain-specific representation, thereby mitigating catastrophic forgetting. To quantify the resulting increase in model complexity, we compute the FLOPs and parameter of each TM. The TM adopts multi-head self-attention (MHSA), a feed-forward network (FFN), residual connections, and layer normalization. A single TM introduces 228.7 M FLOPs and 1.774 M parameters, representing a negligible memory footprint compared to the substantial storage requirements of old exemplars \cite{liu2025Diverse}. This design enables our DSKC to effectively learn and retain domain-specific knowledge, alleviating catastrophic forgetting with minimal computational overhead.
\begin{table}[ht]
	\centering
	\renewcommand{\arraystretch}{1.3}
	\setlength{\tabcolsep}{3.5pt}
	\caption{Experimental results regarding training time, Parameters (Para), and FLOPs.}
	\label{tab:complexity}
	\begin{tabular}{c|c|c|c}
		\hline
		Method &Training time (h-min) &Para(M) &FLOPs(G)\\
		\hline
		DRE &20h-18min &99.83 &21.2 \\ 
		PAEMA &6h-20min &86.03 &11.3  \\
		DCR &19h-10min &129.5 &12.6 \\
		DSKC-IM &6h-35min &102.7 &18.3  \\
		Ours &3h-22min &39.67 &8.1  \\
		\hline  
	\end{tabular}
\end{table}\\
\textbf{Efficiency of Computational Complexity: } We conduct experiments to evaluate the computational complexity of DSKC-IM in comparison with state-of-the-art ViT-based methods, including DRE \cite{liu2025Diverse}, PAEMA \cite{li2024exemplar}, and DCR \cite{liu2025domain}, as summarized in Table \ref{tab:complexity}. Training efficiency is reported in terms of hours (h) and minutes (min). Although DSKC-IM entails higher FLOPs and parameter counts than DRE, DCR, and PAEMA, it achieves a notably shorter training duration than DRE and DCR. More importantly, DSKC-IM delivers significantly superior performance, yielding improvements of at least 1.5\%/0.7\% and 2.1\%/2.6\% (mAP/R-1) in Seen-Avg, as well as \textbf{1.8\%}/\textbf{2.2\%} and \textbf{2.6\%}/\textbf{2.2\%} (mAP/R@1) in Unseen-Avg across the two training orders (see Tables \ref{tab:Table2} and \ref{tab:Table3}). To optimize efficiency, our standard DSKC utilizes the ViT-Small/16 \cite{dosovitskiy2020image} architecture as the backbone. Our core strategies (UKL, UKA, and DST) with a rehearsal-free and distillation-free framework are highly compatible with this backbone, substantially enhancing both anti-forgetting and generalization capabilities.
\subsection{Visualization} 
To further validate the effectiveness of our DSKC, we conduct a qualitative comparison with the DASK method, as shown in Figure. \ref{fig:visual}. The left column displays challenging query images, including cases with low-light, occlusion, back views, and side views. In Figure. \ref{fig:visual}, correctly matched pedestrians are marked with green boxes, while incorrect matches are highlighted in red. Under low-light conditions (first row), our DSKC accurately retrieves eight out of ten individuals with the same identity, despite the limited visual information available. Under occlusion conditions (second and third rows), the DASK method matches at least four people incorrectly among ten individuals. But our DSKC accurately retrieves nine out of ten individuals with the same identity. In back-view scenarios (fourth and fifth rows), our DSKC successfully handles angle change issue, outperforming DASK method in matching the same identity. For side-view (sixth row), the DASK method only correctly matches four pedestrians, whereas our DSKC achieves perfect retrieval (10/10) pedestrians with the occlusion and side view. These results demonstrate that our DSKC achieves superior retrieval performance, consistently identifying more pedestrians with the same identity across diverse challenging conditions. This advantage stems from our DSKC, which effectively consolidates the unified representation with the characteristics of cross-domain sharing and identity discrimination to enhance the model’s anti-forgetting and matching accuracy performance.\\
\section{Conclusions}
We propose DSKC, a novel rehearsal-free and distillation-free framework. To consolidate diverse domain styles into unified knowledge, we develop a domain-style encoder and a unified knowledge learning module, which capture instance-level domain-specific representations and adaptively integrate them into a unified space. Furthermore, we introduce unified knowledge association and domain-based style transfer to align domain-specific representation with the unified representation through instance-wise correlation and domain-wise alignment. Extensive experiments demonstrate that our DSKC achieves superior anti-forgetting performance and generalization capability.

\bibliographystyle{IEEEtran}
\bibliography{DSKC.bib}

\vspace{11pt}
\vspace{-33pt}
\begin{IEEEbiography}[{\includegraphics[width=1in,height=1.25in,clip,keepaspectratio]{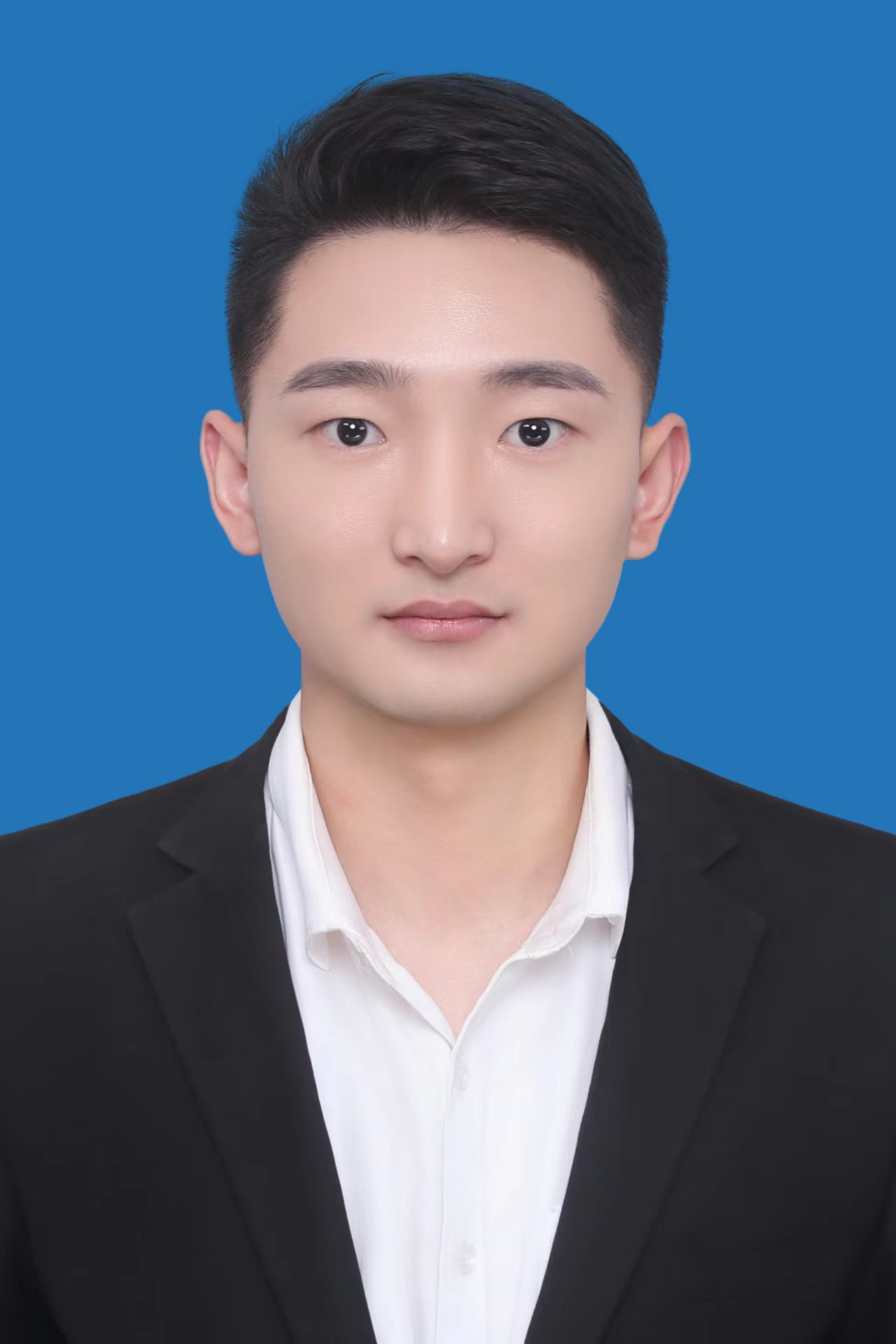}}]{Shiben Liu} received his B.E. and M.S. degrees in Electronic Information Engineering, and Communication and Information Systems from Liaoning University of Engineering and Technology, China, in 2019 and 2022, respectively. He is currently pursuing a Ph.D. degree in State Key Laboratory of Robotics, Shenyang Institute of Automation, University of Chinese Academy of Sciences, Beijing, China. His current research focuses on deep learning, lifelong learning, person re-identification, image restoration and analysis. He has served	as the Reviewer for the international journals such as the TNNLS, TCSVT, TMM, KBS, RAL, Neurocomputing, and so on.	
%IEEE TRANSACTIONS ON CIRCUITS AND SYSTEMS FOR VIDEO TECHNOLOGY, IEEE TRANSACTIONS ON MULTIMEDIA, IEEE ROBOTICS AND AUTOMATION LETTERS, NEUROCOMPUTING, and so on.
\end{IEEEbiography}
\vspace{11pt}
\vspace{-33pt}
\begin{IEEEbiography}[{\includegraphics[width=1in,height=1.25in,clip,keepaspectratio]{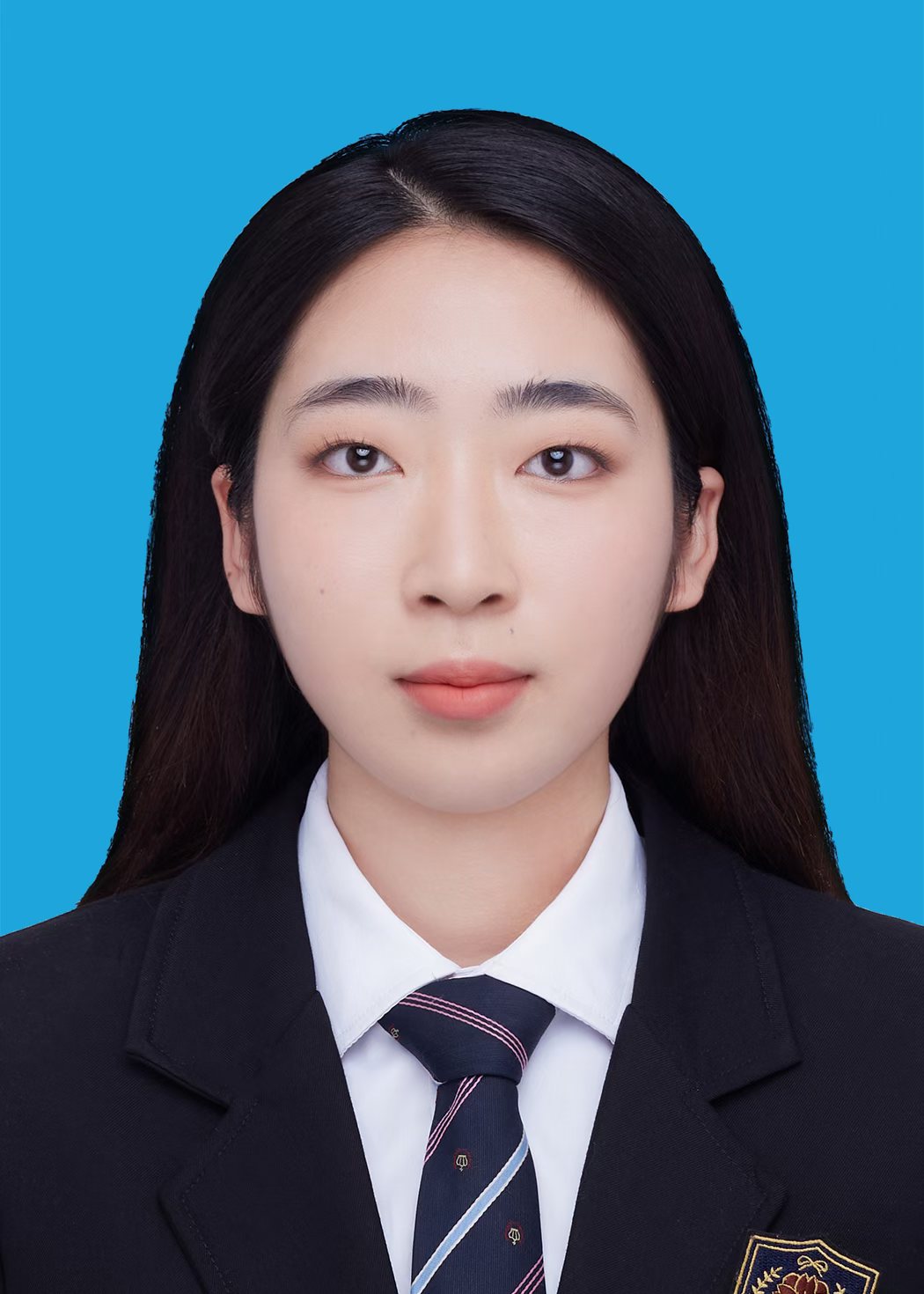}}]{Mingyue Xu} received her B.E. degree in Information Security from Guizhou University, China, in 2024. She is currently pursuing an M.S. degree at the State Key Laboratory of Robotics, Shenyang Institute of Automation, University of Chinese Academy of Sciences. Her current research focuses on deep learning, lifelong learning, open vocabulary, and multiple object tracking.
\end{IEEEbiography}
\vspace{11pt}
\vspace{-33pt}
\begin{IEEEbiography}[{\includegraphics[width=1in,height=1.25in,clip,keepaspectratio]{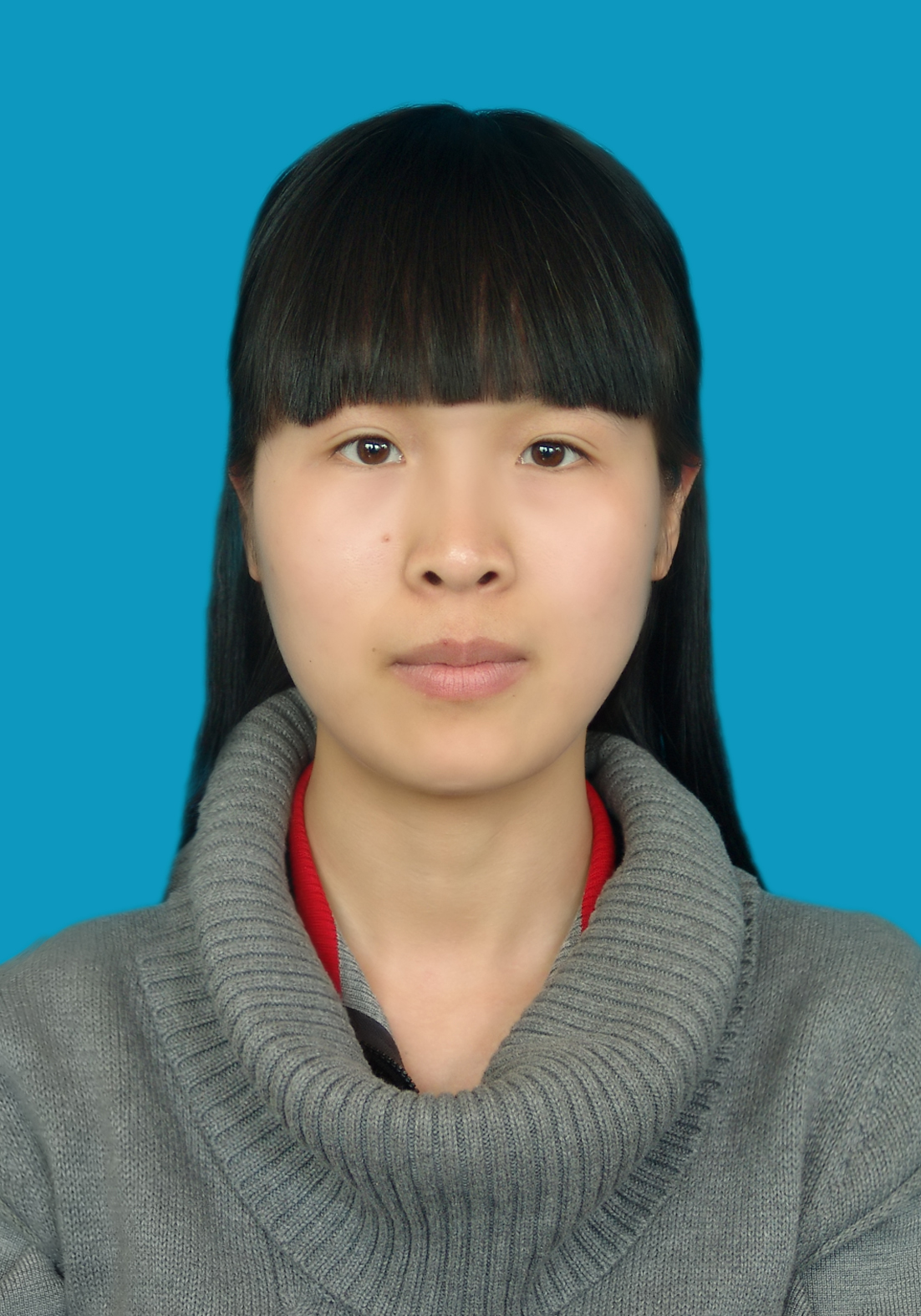}}]{Huijie Fan} (Member, IEEE) received the B.E. degree in automation from the University of Science and Technology of Science and Technology of China, China, in 2007, and the Ph.D. degree in mode recognition and intelligent systems from the Chinese Academy of Sciences University, Beijing, China, in 2014. She is currently a Research Scientist with the Institute of Shenyang Automation of the Chinese Academy of Sciences. Her research interests include deep learning on image processing and medical image processing and applications.
\end{IEEEbiography}
\vspace{11pt}
\vspace{-33pt}
\begin{IEEEbiography}[{\includegraphics[width=1in,height=1.25in,clip,keepaspectratio]{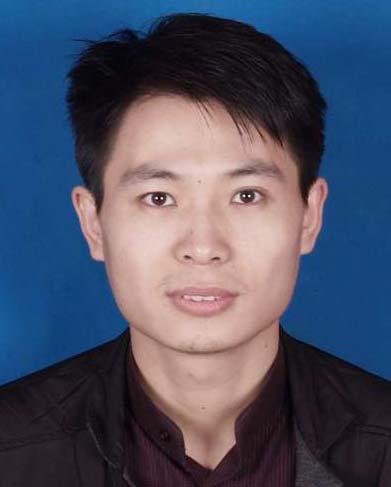}}]{Qiang Wang} received the B.E. and M.S. degrees in school of Computer Science and Technology from Shandong Jianzhu University and Tianjin Normal University, P.R.China,in 2004 and 2008, respectively, the Ph.D degree in State Key Laboratory of Robotics, Shenyang Institute of Automation, University of Chinese Academy of Sciences, Beijing, China in 2020. He is an Associate Professor in the Key Laboratory of Manufacturing Industrial Integrated in Shenyang University. He has some top-tier journal papers accepted at TIP, TMM, TCSVT, IoT-J and Pattern Recognition et al. His current research focuses on deep learning, multi-task learning, image restoration and analysis.
\end{IEEEbiography}
\vspace{11pt}
\vspace{-33pt}
\begin{IEEEbiography}[{\includegraphics[width=1in,height=1.25in,clip,keepaspectratio]{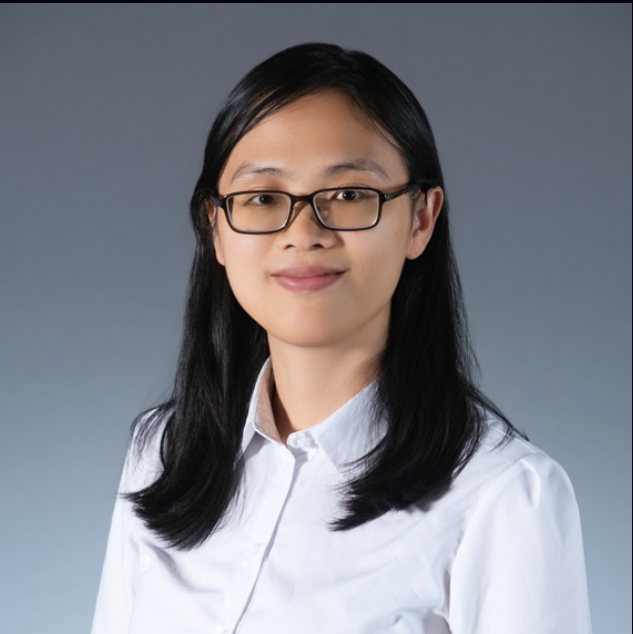}}]{Liangqiong Qu} received the Ph.D.degree in pattern recognition and intelligent system from the Uni- versity of ChineseAcademy of Sciences, Beijing, China, in 2018, and the Ph.D.degree in computer science from the City University of Hong Kong, Hong Kong,in 2018.She was a Post-Doctoral Re- search Fellow with The University of North Carolina at Chapel Hill, Chapel Hill, NC, USA,from 2018 to 2019, and a Post-Doctoral Research Fellow with Stanford University, Stanford, CA, USA.She is cur- rently an Assistant Professor with the Department of Statistics and Actuarial Science, The University of Hong Kong.Her research interests include computer vision, machine learning, and deep learning, with a focus on applications to health care.
\end{IEEEbiography}
\vspace{11pt}
\vspace{-33pt}
\begin{IEEEbiography}[{\includegraphics[width=1in,height=1.25in,clip,keepaspectratio]{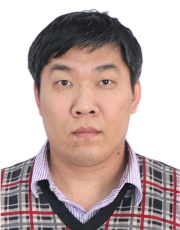}}]{Zhi Han} (Member, IEEE) received the B.S. and M.S. degrees in applied mathematics from Xi’an Jiaotong University (XJTU), Xi’an, China, in 2005 and 2007, respectively, the joint Ph.D. degree in statistics from the University of California at Los Angeles (UCLA),Los Angeles, CA, USA, in 2011, and the Ph.D.degree in applied mathematics from Xi’an Jiaotong University (XJTU) in 2012.He is currently a Professor with the State Key Laboratory of Robotics, Shenyang Institute of Automation (SIA), Chinese Academy of Sciences (CAS). His research interests include image/video modeling, low-rank matrix recovery, and deep neural networks.
\end{IEEEbiography}
\vspace{11pt}
\vspace{-33pt}
\vfill
\end{document}